\newif\iftechreport 
\techreporttrue

\iftechreport
\documentclass[a4paper]{article}
\usepackage{mythm}
\newenvironment{keywords}{\noindent{\bf Keywords.}}{}
\else
\documentclass[a4paper,oribibl,envcountsame]{llncs}
\fi

\usepackage[english]{babel}
\usepackage[utf8]{inputenc}
\usepackage[bookmarks]{hyperref}
\usepackage{enumerate}
\usepackage{amsmath}
\usepackage{amssymb}
\usepackage{paralist}
\usepackage{calc}

\usepackage{tikz}
\usetikzlibrary{shapes,arrows}

\usepackage{pgfplots}
\usepackage{pgfplotstable}
\pgfplotsset{compat=newest}

\usepackage[labelfont=bf]{caption}
\newcommand*\samethanks[1][\value{footnote}]{\footnotemark[#1]}
\AtBeginDocument{

}
\newcommand{\appendixref}[1]{{\hyperref[#1]{Appendix~\ref*{#1}}}}

\usepackage{wasysym}
\newcommand{\happy}{\mbox{\rm \smiley{}}} 
\def\E{\mathbb{E}}  
\def\A{\mathcal{A}} 
\def\O{\mathcal{O}} 

\title{A Definition of Happiness \texorpdfstring{\\}{} for Reinforcement Learning Agents%
\iftechreport\thanks{%
Research supported by the People for the Ethical Treatment of Reinforcement Learners
\url{http://petrl.org}.
This is the extended technical report.
The final publication is available at \url{http://link.springer.com}.
}\else\thanks{%
Research supported by the People for the Ethical Treatment of Reinforcement Learners
\url{http://petrl.org}.
See the extended technical report
for omitted proofs and details about the data analysis~\cite{DL:2015happinessx}.
}\fi}
\author{Mayank Daswani\thanks{Both authors contributed equally.}
	\and Jan Leike\samethanks}
\iftechreport\else
\titlerunning{A Definition of Happiness for Reinforcement Learning Agents}
\authorrunning{M.\ Daswani and J.\ Leike}
\institute{
  Australian National University \\
  \texttt{\{mayank.daswani|jan.leike\}@anu.edu.au}
}
\fi
\date{\today}

\begin{document}

\maketitle

\begin{abstract}%
What is happiness for reinforcement learning agents?
We seek a formal definition
satisfying a list of desiderata.
Our proposed definition of happiness is
the \emph{temporal difference error},
i.e.\ the difference between the value of the obtained reward and observation
and the agent's expectation of this value.
This definition satisfies most of our desiderata and
is compatible with empirical research on humans.
We state several implications and discuss examples.
\end{abstract}

\begin{keywords}
Temporal difference error,
reward prediction error,
pleasure,
well-being,
optimism,
machine ethics
\end{keywords}

\section{Introduction}
\label{sec:introduction}

People are constantly in search of better ways to be happy.
However, philosophers and psychologists
have not yet agreed on a notion of human happiness.
In this paper, we pursue the more general goal of defining happiness
for intelligent agents.
%
We focus on the reinforcement learning (RL) setting~\cite{SB:1998}
because it is an intensively studied formal framework
which makes it easier to make precise statements.
Moreover, reinforcement learning has been used to model
behaviour in both human and non-human animals~\cite{Niv:2009}.

Here, we decouple the discussion of happiness from
the discussion of consciousness, experience, or qualia.
We completely disregard whether happiness is actually consciously experienced
or what this means.
The problem of consciousness has to be solved separately; but its answer might matter insofar that
it could tell us which agents' happiness we should care about.

\paragraph{Desiderata.}
We can simply ask a human how happy they are.
But artificial reinforcement learning agents cannot yet speak.
Therefore we use our human
``common sense'' intuitions about
happiness to come up with a definition.
We arrive at the following desired properties.
\begin{itemize}
\item \emph{Scaling.}
	Happiness should be invariant under scaling of the rewards.
	Replacing every reward $r_t$ by
	$c r_t + d$ for some $c, d \in \mathbb{R}$ with $c > 0$
	(independent of $t$)
	does not change the reinforcement learning problem in any relevant way.
	Therefore we desire a happiness measure to be independent
	under rescaling of the rewards.

\item \emph{Subjectivity.}
	Happiness is a subjective property of the agent
	depending only on information available to the agent.
	For example, it cannot depend on the true environment.
\item \emph{Commensurability.}
	The happiness of different agents should be comparable.
	If at some time step an agent $A$ has happiness $x$,
	and another agent $B$ has happiness $y$,
	then it should be possible to tell whether $A$ is happier than $B$
	by computing $x - y$.
	This could be relaxed
	by instead asking that
	$A$ can calculate the happiness of $B$
	\emph{according to $A$'s subjective beliefs.}
\item \emph{Agreement.}
	The happiness function should match experimental data about human happiness.
\end{itemize}

It has to be emphasised that
in humans,
happiness cannot be equated with pleasure~\cite{RSDD:2014}.
In the reinforcement learning setting,
pleasure corresponds to the reward.
Therefore happiness and reward have to be distinguished.
We crudely summarise this as follows;
for a more detailed discussion see \autoref{sec:definition-of-happiness}.
\[
 \text{pleasure} ~=~ \text{reward} ~\neq~ \text{happiness}
\]
The happiness measure that we propose is the following.
An agent's happiness in a time step $t$ is the difference
between the value of the obtained reward and observation
and the agent's expectation of this value at time step $t$.
In the Markov setting,
this is also known as the \emph{temporal difference error} (TD error)~\cite{SB:1990}.
However, we do not limit ourselves to the Markov setting in this paper.
In parts of the mammalian brain,
the neuromodulator dopamine has a strong connection to the TD error~\cite{Niv:2009}.
Note that while our definition of happiness is not equal to reward
it remains highly correlated to the reward,
especially if the expectation of the reward is close to 0.

Our definition of happiness coincides with
the definition for \emph{joy} given by Jacobs et al.~\cite{JBJ:2014},
except that the latter is weighted by
$1$ minus the (objective) probability of taking the transition
which violates subjectivity.
Schmidhuber's work on `intrinsic motivation' adds a related component to the reward
in order to motivate the agent to explore in interesting directions~\cite{Schmidhuber:2010}.

Our definition of happiness can be split into
two parts.
\begin{inparaenum}[(1)]
\item The difference between the instantaneous reward and its expectation,
	which we call \emph{payout}, and
\item how the latest observation
	and reward changes the agent's
	estimate of future rewards, which
	we call \emph{good news}.
\end{inparaenum}
Moreover, we identify two sources of happiness:
\emph{luck}, favourable chance outcomes
(e.g.\ rolling a six on a fair
die), and
\emph{pessimism},
having low expectations of the environment
(e.g.\ expecting a fair die to be
biased against you).
We show that agents that know
the world perfectly have zero expected happiness.
\iftechreport
Proofs can be found in \appendixref{app:omitted-proofs}.
\fi

In the rest of the paper, we use our definition as a starting point
to investigate the following questions.
Is an off-policy agent happier than an on-policy one? Do monotonically
increasing rewards necessarily imply a happy agent? How does value
function initialisation affect the happiness of an agent? Can we
construct an agent that maximises its own happiness?

\section{Reinforcement Learning}
\label{sec:reinforcement-learning}

In reinforcement learning (RL)
an \emph{agent} interacts with an \emph{environment} in cycles:
at time step $t$ the agent chooses an \emph{action} $a_t \in \A$ and
receives an \emph{observation} $o_t \in \O$
and a real-valued \emph{reward} $r_t \in \mathbb{R}$;
the cycle then repeats for time step $t + 1$~\cite{SB:1998}.
The list of interactions $a_1 o_1 r_1 a_2 o_2 r_2 \ldots$
is called a \emph{history}.
We use $h_t$ to denote a history of length $t$,
and we use the shorthand notation $h := h_{t-1}$ and $h' := h_{t-1} a_t o_t r_t$.
The agent's goal is to choose actions to maximise cumulative rewards.
To avoid infinite sums,
we use a \emph{discount factor $\gamma$}
with $0 < \gamma < 1$
and maximise the discounted sum $\sum_{t=1}^\infty \gamma^t r_t$.
A \emph{policy} is a function $\pi$
mapping every history to the action taken after seeing this history, and
an \emph{environment} $\mu$ is a stochastic mapping from histories to
observation-reward-tuples.

A policy $\pi$ together with an environment $\mu$
yields a probability distribution over histories.
Given a random variable $X$ over histories,
we write the $\pi$-$\mu$-expectation of $X$ conditional on the history $h$ as
$\E^\pi_\mu[X \mid h]$.

The \emph{(true) value function} $V^\pi_\mu$
of a policy $\pi$ in environment $\mu$
maps a history $h_t$ to the
expected total future reward
when interacting with environment $\mu$ and
taking actions according to the policy $\pi$:
\begin{equation}\label{eq:value-function}
V^\pi_\mu(h_t) := \E^\pi_\mu \left[\textstyle \sum_{k=t+1}^\infty \gamma^{k-t-1} r_k \mid h_t \right].
\end{equation}
It is important to emphasise that $\E^\pi_\mu$ denotes the \emph{objective}
expectation that can be calculated only by knowing the environment $\mu$.
The \emph{optimal value function} $V^*_\mu$ is defined as the value function
of the optimal policy, $V^*_\mu(h) := \sup_\pi V^\pi_\mu(h)$.

Typically, reinforcement learners do not know the environment and
are trying to learn it.
We model this by assuming that
at every time step the agent has (explicitly or implicitly)
an estimate $\hat{V}$ of the value function $V^\pi_\mu$.
Formally,
a \emph{value function estimator} maps a history $h$ to
a value function estimate $\hat{V}$.
Finally, we define an \emph{agent} to be
a policy together with a value function estimator.
If the history is clear from context,
we refer to the output of the value function estimator as
\emph{the agent's estimated value}.

If $\mu$ only depends on the last observation and action,
$\mu$ is called \emph{Markov decision process (MDP)}.
In this case,
$\mu(o_t r_t \mid h_{t-1} a_t) = \mu(o_t r_t \mid o_{t-1} a_t)$ and
the observations are called states ($s_t = o_t$).
In MDPs we use the $Q$-value function, the value of a state-action pair,
defined as
$Q^\pi_\mu(s_t,a_t) := \E^\pi_\mu \left[\sum_{k=t+1}^\infty \gamma^{k-t-1} r_k \mid s_t a_t \right]$.
Assuming that the environment is an MDP is very common in the RL literature,
but here we will not make this assumption.

\section{A Formal Definition of Happiness}
\label{sec:definition-of-happiness}

The goal of a reinforcement learning agent is to maximise rewards,
so it seems natural to suppose an agent is happier the more rewards it gets.
But this does not conform to our intuition:
sometimes enjoying pleasures just fails to provide happiness, and
reversely, enduring suffering does not necessarily entail unhappiness
(see \autoref{ex:bad-news} and \autoref{ex:increasing-rewards}).
In fact, it has been shown empirically that rewards and happiness
cannot be equated~\cite{RSDD:2014} ($p$-value $< 0.0001$).

There is also a formal problem with defining happiness in terms of reward:
we can add a constant $c \in \mathbb{R}$ to every reward.
No matter how the agent-environment interaction plays out,
the agent will have received additional cumulative rewards
$C := \sum_{i=1}^t c$.
However, this did not change
the structure of the reinforcement learning problem in any way.
Actions that were optimal before are still optimal and
actions that are slightly suboptimal are still slightly suboptimal
\emph{to the same degree}.
For the agent,
no essential difference between the original reinforcement learning problem
and the new problem can be detected:
in a sense the two problems are \emph{isomorphic}.
If we were to define an agent's happiness as received reward,
then an agent's happiness would vary wildly
when we add a constant to the reward
while the problem stays structurally exactly the same.

We propose the following definition of happiness.

\begin{definition}[Happiness]
\label{def:happiness}
The \emph{happiness} of a reinforcement learning agent
with estimated value $\hat{V}$
at time step $t$ with history $ha_t$
while receiving observation $o_t$ and reward $r_t$ is
\begin{equation}\label{eq:happiness}
   \happy(ha_t o_t r_t, \hat{V})
:= r_t + \gamma \hat{V}(ha_t o_t r_t) - \hat{V}(h).
\end{equation}
If $\happy(h', \hat{V})$ is positive,
we say the agent is \emph{happy}, and
if $\happy(h', \hat{V})$ is negative,
we say the agent is \emph{unhappy}.
\end{definition}

It is important to emphasise that $\hat{V}$ represents the agent's
\emph{subjective} estimate of the value function.
If the agent is good at learning,
this might converge to something close to
the true value function $V^\pi_\mu$.
%
In an MDP \eqref{eq:happiness} is also known as
the \emph{temporal difference error}~\cite{SB:1990}.
This number is used used to update the value function,
and thus plays an integral part in learning.

If there exists a probability distribution $\rho$ on histories such that
the value function estimate $\hat{V}$ is given by
the expected future discounted rewards
according to the probability distribution $\rho$,
\begin{equation}\label{eq:subjective-expectation}
  \hat{V}(h)
= \E^\pi_\rho \left[ \textstyle \sum_{k=t+1}^\infty \gamma^{k-t-1} r_k \mid h \right],
\end{equation}
then we call $E := \E^\pi_\rho$ the agent's \emph{subjective expectation}.
Note that we can always find such a probability distribution,
but this notion only really makes sense for
\emph{model-based agents} (agents that learn a model of their environment).
Using the agent's subjective expectation,
we can rewrite \autoref{def:happiness} as follows.

\begin{proposition}[Happiness as Subjective Expectation]
\label{prop:happiness-as-subjective-expectation}
Let $E$ denote an agent's subjective expectation. Then
\begin{equation}\label{eq:happiness-as-subjective-expectation}
  \happy(h', \hat{V})
= r_t - E[r_t \mid h]
  + \gamma \left( \hat{V}(h') - E[\hat{V}(haor) \mid h] \right).
\end{equation}
\end{proposition}

\autoref{prop:happiness-as-subjective-expectation} states that
happiness is given by the difference of
how good the agent thought it was doing and
what it learns about how well it actually does.
We distinguish the following two components in \eqref{eq:happiness-as-subjective-expectation}:
\begin{itemize}
\item \emph{Payout:}
	the difference of the obtained reward $r_t$ and
	the agent's expectation of that reward $E[r_t \mid h]$.
\item \emph{Good News:}
	the change in opinion of the expected future rewards
	after receiving the new information $o_t r_t$.
\end{itemize}
\[
  \happy(h', \hat{V})
= \underbrace{r_t - E[r_t \mid h]}_{\text{payout}}
  + \gamma \big( \underbrace{\hat{V}(h') - E[\hat{V}(haor) \mid h]}_{\text{good news}} \big)
\]

\begin{example}\label{ex:bad-news}
Mary is travelling on an air plane.
She knows that air planes crash very rarely,
and so is completely at ease.
Unfortunately she is flying on a budget airline,
so she has to pay for her food and drink.
A flight attendant comes to her seat and gives
her a free beverage.
Just as she starts drinking it,
the intercom informs everyone that the engines have failed.
Mary feels some happiness from the free drink (\emph{payout}),
but her expected future reward is much lower than in the state before learning
the \emph{bad news}.
Thus overall, Mary is unhappy.
\end{example}

For each of the two components, payout and good news,
we distinguish the following two sources of happiness.
\begin{itemize}
\item \emph{Pessimism:}%
\footnote{%
\emph{Optimism} is a standard term in the RL literature
to denote the opposite phenomenon.
However, this notion is somewhat in discord with optimism in humans.
}
	the agent expects the environment to contain less rewards than it actually does.
\item \emph{Luck:}
	the outcome of $r_t$ is unusually high due to randomness.
\end{itemize}
\vspace{-3mm}
\begin{align*}
   r_t - E[r_t \mid h]
=~ &\underbrace{r_t - \E^\pi_\mu[ r_t \mid h]}_{\text{luck}}
   + \underbrace{\E^\pi_\mu[ r_t \mid h] - E[r_t \mid h]}_{\text{pessimism}} \\
   \hat{V}(h') - E[\hat{V}(haor) \mid h]
=~ &\underbrace{\hat{V}(h') - \E^\pi_\mu[ \hat{V}(haor) \mid h]}_{\text{luck}} \\
   &+ \underbrace{\E^\pi_\mu[ \hat{V}(haor) \mid h] - E[ \hat{V}(haor) \mid h]}_{\text{pessimism}}
\end{align*}
\vspace{-4mm} 

\begin{example}
Suppose Mary fears flying and expected the plane to crash
(\emph{pessimism}).
On hearing that the engines failed (\emph{bad luck}),
Mary does not experience very much change in her future expected reward.
Thus she is happy that she (at least) got a free drink.
\end{example}

The following proposition states that
once an agent has learned the environment,
its expected happiness is zero.
In this case, underestimation cannot contribute to happiness and thus
the only source of happiness is luck, which cancels out in expectation.

\begin{proposition}[Happiness of Informed Agents]
\label{prop:informed-agent}
An agent that knows the world has an expected happiness of zero:
for every policy $\pi$ and every history $h$,
\[
\E^\pi_\mu[\happy(h', V^\pi_\mu) \mid h] = 0.
\]
\end{proposition}

Analogously,
if the environment is deterministic,
then luck cannot be a source of happiness.
In this case, happiness reduces to how much the agent
underestimates the environment.
By \autoref{prop:informed-agent},
having learned a deterministic environment perfectly,
the agent's happiness is equal to zero.

\section{Matching the Desiderata}
\label{sec:matching-the-desiderata}

Here we discuss in which sense our definition of happiness
satisfies the desiderata from \autoref{sec:introduction}.

\paragraph{Scaling.}
If we transform the rewards to $r'_t = cr_t + d$ with $c > 0$, $d \in \mathbb{R}$
for each time step $t$
without changing the value function,
the value of $\happy$ will be completely different.
However, a sensible learning algorithm should be able to
adapt to the new reinforcement learning problem with the scaled rewards
without too much problem.
At that point, the value function gets scaled as well,
$V_{new}(h) = c V(h) + d/(1 - \gamma)$.
In this case we get
\begin{align*}
   \happy(ha_t o_t r', V_{new})
&= r'_t + \gamma V_{new}(ha_t o_t r'_t) - V_{new}(h) \\
&= cr_t + d + \gamma c V(ha_t o_t r'_t) + \gamma \frac{d}{1 - \gamma} - c V(h) - \frac{d}{1 - \gamma} \\
&= c \big( r_t + \gamma V(ha_t o_t r'_t) - V(h) \big),
\end{align*}
hence happiness gets scaled by a positive factor
and thus its sign remains the same,
which would not hold if we defined happiness just in terms of rewards.

\paragraph{Subjectivity.}
The definition \eqref{eq:happiness-as-subjective-expectation} of $\happy$
depends only on the current reward and
the agent's current estimation of the value function,
both of which are available to the agent.

\paragraph{Commensurability.}
The scaling property as described above means that the
exact value of the happiness is not useful in comparing two agents,
but the sign of the total happiness can at least tell us whether a
given agent is happy or unhappy.
Arguably,
failing this desideratum is not surprising;
in utility theory the utilities/rewards
of different agents are typically not commensurable either.

However,
given two agents $A$ and $B$,
$A$ can still calculate the $A$-subjective
happiness of a history experienced by $B$
as $\happy({haor}_B,\hat{V}^A)$.
This corresponds to the human intuition of
``putting yourself in someone else's shoes''.
If both agents are acting in the
same environment, the resulting numbers should be commensurable,
since the calculation is done using the same value function.
It is entirely possible that $A$ believes $B$
to be happier, i.e.\
$\happy({haor}_B,\hat{V}^A) > \happy({haor}_A,\hat{V}^A)$,
but also that $B$ believes $A$ to be happier
$\happy({haor}_A,\hat{V}^B) > \happy({haor}_B,\hat{V}^B)$,
because they have different expectations of the environment.

\paragraph{Agreement.}
Rutledge et al.\ measure subjective well-being
on a smartphone-based experiment with 18,420 participants~\cite{RSDD:2014}.
In the experiment, a subject goes through 30 trials in each of which
they can choose between a sure reward and
a gamble that is resolved within a short delay.
Every two to three trials the subjects are asked to
indicate their momentary happiness.

Our model based on \autoref{prop:happiness-as-subjective-expectation}
with a very simple learning algorithm and no loss aversion
correlates fairly well with reported happiness
(mean $r = 0.56$, median $r^2 = 0.41$, median $R^2 = 0.27$)
while fitting individual discount factors,
comparative to Rutledge et al.'s model
(mean $r = 0.60$, median $r^2 = 0.47$, median $R^2 = 0.36$) and
a happiness=cumulative reward model
(mean $r = 0.59$, median $r^2 = 0.46$, median $R^2 = 0.35$).
This analysis is inconclusive, but unsurprisingly so:
the expected reward is close to $0$ and thus
our happiness model correlates well with rewards.
\iftechreport
See \appendixref{app:data-analysis} for the details of our data analysis.
\else\fi

The \emph{hedonic treadmill}~\cite{Brickman:1971} refers
to the idea that humans return to a \emph{baseline level
of happiness} after significant negative or positive events. Studies
have looked at lottery winners and accident victims~\cite{Brickman:1978},
and people dealing with
paralysis, marriage, divorce, having children
and other life changes~\cite{Diener:2006}. In most cases
these studies have observed a return to baseline happiness after
some period of time has passed;
people learn to make correct reward predictions again.
Hence their expected happiness returns to zero (\autoref{prop:informed-agent}).
Our definition unfortunately does not explain why people have different
baseline levels of happiness (or hedonic set points), but these may be
perhaps explained by biological means (different humans have different
levels of neuromodulators, neurotransmitters, hormones, etc.) which may move
their baseline happiness.
Alternatively, people might simply learn to associate
different levels of happiness with ``feeling happy'' according to their
environment.

\section{Discussion and Examples}
\label{sec:discussion-and-examples}

\subsection{Off-policy Agents}

In reinforcement learning, we are mostly interested in learning the
value function of the optimal policy. A common difference between RL
algorithms is whether they learn \emph{off-policy} or \emph{on-policy}.
An \emph{on-policy} agent evaluates the value of the policy it is currently following.
For example, the policy that the agent is made to follow could be an
$\varepsilon$-greedy policy,
where the agent picks $\arg\max_{a}Q^{\pi}(h,a)$ a fraction
$(1-\varepsilon)$ of the time, and a random action otherwise.
If $\varepsilon$ is decreased to zero over time,
then the agent's learned policy tends to the optimal policy in MDPs.
Alternatively, an agent can learn \emph{off-policy}, that is it can
learn about one policy (say, the optimal one) while following a different
\emph{behaviour policy}.

The behaviour policy ($\pi_{b}$) determines how the agent
acts while it is learning the optimal policy.
Once an off-policy learning agent
has learned the optimal value function $V^*_\mu$,
then it is not happy
if it still acts according to some other (possibly suboptimal) policy.

\begin{proposition}[Happiness of Off-Policy Learning]
\label{prop:off-policy-learning}
Let $\pi$ be some policy and $\mu$ be some environment.
Then for any history $h$
\[
\E^\pi_\mu[\happy(h', V^*_\mu) \mid h] \leq 0.
\]
\end{proposition}

Q-learning is an example of an off-policy
algorithm in the MDP setting.
If Q-learning converges, and the agent is still following
the sub-optimal behaviour policy then
\autoref{prop:off-policy-learning} tells us that
the agent will be unhappy.
Moreover, this means that SARSA (an on-policy RL algorithm)
will be happier than Q-learning on average and in expectation.

\subsection{Increasing and Decreasing Rewards}

Intuitively, it seems that if things are constantly getting better,
this should increase happiness.
However, this is not generally the case:
even an agent that obtains monotonically increasing rewards
can be unhappy if it thinks that these rewards mean
even higher negative rewards in the future.

\begin{example}\label{ex:increasing-rewards}
Alice has signed up for a questionable drug trial which
examines the effects of a potentially harmful drug.
This drug causes temporary pleasure to the user every time it is used,
and increased usage results in increased pleasure.
However, the drug reduces quality of life in the long term.
Alice has been informed of the potential side-effects of the drug.
She can be either part of a placebo group or the group given the drug.
Every morning Alice is given an injection of an unknown liquid.
She finds herself feeling temporary
but intense feelings of pleasure.
This is evidence  that she is in the non-placebo group,
and thus has a potentially reduced quality of life in the long term.
Even though she experiences pleasure (increasing rewards)
it is evidence of very \emph{bad news} and thus she is unhappy.
\end{example}

Analogously, decreasing rewards do not generally imply unhappiness.
For example, the pains of hard labour can mean happiness
if one expects to harvest the fruits of this labour in the future.

\subsection{Value Function Initialisation}

\newlength{\figlength}
\begin{figure}[t]
\centering
\begin{minipage}{.47\textwidth}
\centering
\tikzstyle{mcirc} = [draw, circle, fill=none]
\tikzstyle{line} = [draw, -latex',font=\scriptsize]
\begin{tikzpicture}
\node [mcirc] (s0) at (0,0) {$s_0$};
\node [mcirc] (s1) at (3,0) {$s_1$};

\path [line] (s0)  edge [bend left=50]  node [midway, above] {$\beta:-1$} (s1);
\path [line] (s1)  edge [bend left=50] node [midway, below] {$\beta:-1$} (s0);
\path [line] (s0) edge[loop above]  node {$\alpha:0$} (s0);
\path [line] (s1) edge[loop above] node {$\alpha:2$} (s1);
\end{tikzpicture}
\vspace{5.6em}
\captionof{figure}{
	MDP of \autoref{ex:pessimism-is-not-easy}
	with transitions labelled with actions $\alpha$ or $\beta$ and rewards.
	We use the discount factor $\gamma=0.5$.
	The agent starts in $s_{0}$.
	Define $\pi_{0}(s_{0}) := \alpha$,
	then $V^{\pi_{0}}(s_{0})=0$.
	The optimal policy is $\pi^{*}(s_{0}) = \beta$,
	so $V^{\pi^{*}}(s_{0})=1$
	and $V^{\pi^{*}}(s_{1})=4$.
}
\label{fig:value-initialization-MDP}
\end{minipage}~~~~
\begin{minipage}{.47\textwidth}
\setlength{\figlength}{\textwidth-\heightof{R}-\heightof{R}-0.85cm}
\begin{tikzpicture}
 \begin{axis}[
axis y line*=left,
xmin =-1, xmax=101,
ymin =-3.5, ymax=2.5,
	legend pos=south west,
	xlabel={Time step},
	ylabel={Happiness},
	y label style={outer sep=5pt,inner sep=0pt,at={(0,0.5)},yshift=0.1cm},
	scale only axis,
	width = \figlength,
      ]
      \addplot+ [no marks, blue]
	plot table[x index=0,y index=1]{data/optimism_ex1.dat};
      \addplot+ [no marks, dashed, black]
	plot table[x index=0,y index=1]{data/pessimism_ex1.dat};
      \addplot+ [no marks,black,densely dotted,red]
	plot table[x index=0,y index=1]{data/optimism_ex1_reward.dat};
\legend{Optimistic , Pessimistic, Opt. (rewards)}
\end{axis}
\begin{axis} [
xmin =-1, xmax=101,
ymin =-3.5, ymax=2.5,
axis y line*=right,
    axis x line=none,
scale only axis,
	width = \figlength,
y label style={outer sep=0pt,inner sep=0pt,yshift=0.3cm},
	ylabel={Rewards},
	]
\end{axis}
\end{tikzpicture}

\captionof{figure}{
	A plot of happiness for \autoref{ex:pessimism-is-not-easy}.
	We use the learning rate $\alpha=0.1$.
	The pessimistic agent has zero happiness (and rewards),
	whereas the optimistic agent is initially unhappy,
	but once it transitions to state $s_1$ becomes happy.
	The plot also shows the rewards of the optimistic agent.
}
\label{fig:value-initialization-plot}
\end{minipage}
\vspace{-5mm} 
\end{figure}

\begin{example}[Increasing Pessimism Does Not Increase Happiness]\label{ex:pessimism-is-not-easy}
Consider the deterministic MDP example in \autoref{fig:value-initialization-MDP}.
Assume that the agent has an initial value function $\hat{Q}_{0}(s_{0},\alpha)=0$,
$\hat{Q}_{0}(s_{0},\beta)=-\varepsilon,\hat{Q}_{0}(s_{1},\alpha)=\varepsilon\mbox{ and }\hat{Q}_{0}(s_{1},\beta)=0$.
If no forced exploration is carried out by the agent, it has no incentive
to visit $s_{1}$. The happiness achieved by such an agent for some
time step $t$ is $\happy(s_{0}\alpha s_{0}0,\hat{V}_{0})=0$
where $\hat{V}_{0}(s_0) := \hat{Q}_0(s_0,\alpha) = 0$.
However, suppose the agent is (more optimistically) initialised
with $\hat{Q}_{0}(s_{0},\alpha)=0,\hat{Q}_{0}(s_{0},\beta)=\varepsilon$.
In this case, the agent would take action $\beta$ and arrive in state
$s_{1}$. This transition would have happiness
$\happy(s_{0}\beta s_{1}{-1},\hat{V}_{0})=-1+\gamma\hat{Q}_{0}(s_{1},\alpha)-\hat{Q}_{0}(s_{0},\beta)=-1-0.5\varepsilon$.
However, the next transition is $s_{1}\alpha s_{1}2$ which has happiness
$\happy(s_{1}\alpha s_{1}2,\hat{V}_{0})=2+\gamma\hat{Q}_{0}(s_{1},\alpha)-\hat{Q}_{0}(s_{1},\alpha)=2-0.5\varepsilon.$
If $\hat{Q}_{0}$ is not updated by some learning mechanism the
agent will continue to accrue this positive happiness for all future
time steps. If the agent does learn, it will still be some time steps
before $\hat{Q}$ converges to $Q^{*}$ and the positive happiness
becomes zero (see \autoref{fig:value-initialization-plot}).
It is arguable whether this agent which suffered one
time step of unhappiness but potentially many time steps of happiness
is overall a happier agent, but it is some evidence
that absolute pessimism does not necessarily lead to the happiest
agents.
\end{example}

\subsection{Maximising Happiness}

How can an agent increase their own happiness?
The first source of happiness, luck,
depends entirely on the outcome of a random event
that the agent has no control over.
However, the agent could modify its learning algorithm to
be systematically pessimistic about the environment.
For example, when fixing the value function estimation
below $r_{\text{min}} / (1 - \gamma)$ for all histories,
happiness is positive at every time step.
But this agent would not actually take any sensible actions.
Just as optimism is commonly used to artificially increase exploration,
pessimism discourages exploration
which leads to poor performance.
As demonstrated in \autoref{ex:pessimism-is-not-easy},
a pessimistic agent may be less happy than a more optimistic one.

Additionally, an agent that explicitly tries to maximise its own happiness
is no longer a reinforcement learner.
So instead of asking how an agent can increase its own happiness,
we should fix a reinforcement learning algorithm
and ask for the environment that would make this algorithm happy.

\section{Conclusion}
\label{sec:conclusion}

An artificial superintelligence might contain subroutines that are capable of suffering,
a phenomenon that Bostrom calls \emph{mind crime}~\cite[Ch.\ 8]{Bostrom:2014}.
More generally, Tomasik argues that even current
reinforcement learning agents
could have moral weight~\cite{Tomasik:2014}.
If this is the case,
then a general theory of happiness for reinforcement learners is essential;
it would enable us to derive ethical standards in the treatment of algorithms.
Our theory is very preliminary and should be thought of
as a small step in this direction.
Many questions are left unanswered, and
we hope to see more research on the suffering of AI agents in the future.

\paragraph{Acknowledgements.}
We thank Marcus Hutter and Brian Tomasik for careful reading
and detailed feedback.
The data from the smartphone experiment was kindly provided by Robb Rutledge.
We are also grateful to many of our friends
for encouragement and interesting discussions.

\bibliographystyle{alpha}
\bibliography{references}

\begin{thebibliography}{RSDD14}

\bibitem[BC71]{Brickman:1971}
Philip Brickman and Donald~T Campbell.
\newblock Hedonic relativism and planning the good society.
\newblock {\em Adaptation-Level Theory}, pages 287--305, 1971.

\bibitem[BCJB78]{Brickman:1978}
Philip Brickman, Dan Coates, and Ronnie Janoff-Bulman.
\newblock Lottery winners and accident victims: Is happiness relative?
\newblock {\em Journal of Personality and Social Psychology}, 36(8):917, 1978.

\bibitem[Bos14]{Bostrom:2014}
Nick Bostrom.
\newblock {\em Superintelligence: Paths, Dangers, Strategies}.
\newblock Oxford University Press, 2014.

\bibitem[DLS06]{Diener:2006}
Ed~Diener, Richard~E Lucas, and Christie~Napa Scollon.
\newblock Beyond the hedonic treadmill: Revising the adaptation theory of
  well-being.
\newblock {\em American Psychologist}, 61(4):305, 2006.

\bibitem[JBJ14]{JBJ:2014}
Elmer Jacobs, Joost Broekens, and Catholijn Jonker.
\newblock Joy, distress, hope, and fear in reinforcement learning.
\newblock In {\em Conference on Autonomous Agents and Multiagent Systems},
  pages 1615--1616, 2014.

\bibitem[Niv09]{Niv:2009}
Yael Niv.
\newblock Reinforcement learning in the brain.
\newblock {\em Journal of Mathematical Psychology}, 53(3):139--154, 2009.

\bibitem[RSDD14]{RSDD:2014}
Robb~B Rutledge, Nikolina Skandali, Peter Dayan, and Raymond~J Dolan.
\newblock A computational and neural model of momentary subjective well-being.
\newblock {\em Proceedings of the National Academy of Sciences}, 2014.

\bibitem[SB90]{SB:1990}
Richard Sutton and Andrew Barto.
\newblock Time-derivative models of {P}avlovian reinforcement.
\newblock In {\em Learning and Computational Neuroscience: Foundations of
  Adaptive Networks}, pages 497--537. MIT Press, 1990.

\bibitem[SB98]{SB:1998}
Richard~S Sutton and Andrew~G Barto.
\newblock {\em Reinforcement Learning: An Introduction}.
\newblock MIT Press, Cambridge, MA, 1998.

\bibitem[Sch10]{Schmidhuber:2010}
Jürgen Schmidhuber.
\newblock Formal theory of creativity, fun, and intrinsic motivation
  (1990--2010).
\newblock {\em IEEE Transactions on Autonomous Mental Development},
  2(3):230--247, 2010.

\bibitem[Tom14]{Tomasik:2014}
Brian Tomasik.
\newblock Do artificial reinforcement-learning agents matter morally?
\newblock Technical report, 2014.
\newblock \url{http://arxiv.org/abs/1410.8233}.

\end{thebibliography}

\iftechreport
\newpage
\appendix

\section{Omitted Proofs}
\label{app:omitted-proofs}

\begin{proof}[\iftechreport Proof \else\fi%
of \autoref{prop:happiness-as-subjective-expectation}]
\begingroup
\allowdisplaybreaks
\begin{align*}
   \happy(h', \hat{V})
&= r_t + \gamma \hat{V}(h') - \hat{V}(h) \\
&= r_t + \gamma \hat{V}(h')
       - E \left[ \sum_{k=t}^\infty \gamma^{k-t} r_k \mid h \right] \\
&= r_t - E[ r_t \mid h]
       + \gamma \hat{V}(h')
       - E \left[ \sum_{k=t+1}^\infty \gamma^{k-t} r_k \mid h \right] \\
&= r_t - E[ r_t \mid h]
       + \gamma \hat{V}(h')
       - \gamma E \left[ E \left[ \sum_{k=t+1}^\infty \gamma^{k-t-1} r_k \mid haor \right] \mid h \right] \\
&= r_t - E[ r_t \mid h]
       + \gamma \hat{V}(h')
       - \gamma E \left[ \hat{V}(haor) \mid h \right],
\end{align*}
\endgroup
where the second to last equality uses
the tower property for conditional expectations.
\iftechreport\else\qed\fi
\end{proof}

\begin{proof}[\iftechreport Proof \else\fi%
of \autoref{prop:informed-agent}]
For the true value function $V^\pi_\mu$,
a subjective expectation exists by definition since
\eqref{eq:subjective-expectation} is the same as \eqref{eq:value-function}.
Hence we can apply \autoref{prop:happiness-as-subjective-expectation}:
\begingroup
\allowdisplaybreaks
\begin{align*}
   &\E^\pi_\mu[\happy(h', V^\pi_\mu) \mid h] \\
=~ &\E^\pi_\mu[ r_t - \E^\pi_\mu[r_t \mid h]
                 + \gamma V^\pi_\mu(h')
                 - \gamma \E^\pi_\mu[ V^\pi_\mu(haor) \mid h]
               \mid h] \\
=~ &\E^\pi_\mu[ r_t \mid h ] - \E^\pi_\mu[ \E^\pi_\mu[r_t \mid h] \mid h ]
   + \gamma \E^\pi_\mu[ V^\pi_\mu(h') \mid h]
   - \gamma \E^\pi_\mu[ \E^\pi_\mu[ V^\pi_\mu(haor) \mid h] \mid h] \\
=~ &\E^\pi_\mu[ r_t \mid h ] - \E^\pi_\mu[r_t \mid h]
   + \gamma \E^\pi_\mu[ V^\pi_\mu(h') \mid h]
   - \gamma \E^\pi_\mu[ V^\pi_\mu(haor) \mid h]
=~ 0
\iftechreport\tag*{\qedhere}\else\tag*{\qed}\fi
\end{align*}
\endgroup
\end{proof}

\begin{proof}[\iftechreport Proof \else\fi%
of \autoref{prop:off-policy-learning}]
Let $h'$ be any history of length $t$, and
let $\pi^*$ denote an optimal policy for environment $\mu$,
i.e., $V^*_\mu = V^{\pi^*}_\mu$.
In this case, we have $V^*_\mu(h) \geq V^*_\mu(h \pi(h))$, and hence
\begin{equation}\label{eq:prop:off-policy-learning}
     \E^{\pi^*}_\mu \left[ \sum_{k=t}^\infty \gamma^{k-t} r_k \mid h \right]
\geq \E^{\pi^*}_\mu \left[ \sum_{k=t}^\infty \gamma^{k-t} r_k \mid h \pi(h) \right].
\end{equation}
We use this in the following:
\begingroup
\allowdisplaybreaks
\begin{align*}
   &\E^\pi_\mu[ \happy(h', V^*_\mu) \mid h ] \\
=~ &\E^\pi_\mu[ r_t + \gamma V^*_\mu(h') - V^*_\mu(h) \mid h ] \\
=~ &\E^\pi_\mu \left[ r_t + \gamma \E^{\pi^*}_\mu \left[ \sum_{k=t+1}^\infty \gamma^{k-t-1} r_k \mid h' \right]
                           - \E^{\pi^*}_\mu \left[ \sum_{k=t}^\infty \gamma^{k-t} r_k \mid h \right]
                \mid h \right] \\
\stackrel{\eqref{eq:prop:off-policy-learning}}{\leq}~
   &\E^\pi_\mu \left[ r_t + \E^{\pi^*}_\mu \left[ \sum_{k=t+1}^\infty \gamma^{k-t} r_k \mid h' \right]
                       - \E^{\pi^*}_\mu \left[ \sum_{k=t}^\infty \gamma^{k-t} r_k \mid h a_t \right]
                \mid h \right] \\
=~ &\E^\pi_\mu \Big[ \underbrace{r_t - \E^{\pi^*}_\mu[r_t \mid ha_t]}_{=0} \\
   &\qquad\qquad\qquad+ \underbrace{\E^{\pi^*}_\mu \left[ \sum_{k=t+1}^\infty \gamma^{k-t} r_k \mid h' \right]
                      - \E^{\pi^*}_\mu \left[ \sum_{k=t+1}^\infty \gamma^{k-t} r_k \mid ha_t \right]}_{=0}
               \mid h \Big]
\end{align*}
\endgroup
since
$
  \E^\pi_\mu[ \E^{\pi^*}_\mu [ X \mid ha_t o_t r_t ] \mid h ]
= \E^\pi_\mu[ \E^{\pi^*}_\mu [ X \mid ha_t ] \mid h]
$
because
conditional on $ha_t$,
the distribution of $o_t$ and $r_t$ is independent of the policy.
\iftechreport\else\qed\fi
\end{proof}

\section{Data Analysis}
\label{app:data-analysis}

In this section we describe the details of the data analysis
on the Great Brain Experiment%
\footnote{\url{http://www.thegreatbrainexperiment.com/}}
conducted by Rutledge et al.~\cite{RSDD:2014}.
This experiment measures subjective well-being
on a smartphone app and had 18,420 subjects.
Each subject goes through 30 trials and starts with 500 points.
At the start of a trial, the subject is given an
option to choose between a certain reward (CR) and a 50--50 gamble between
two outcomes.
The gamble is pictorially shown to have an equal probability
between the outcomes, thus making it easy for the subject
to choose between the certain reward and the gamble.

Before the trials start the agent is asked to rate their
current happiness on a scale of 0-100. The slider
for the scale is initialised randomly.
Every two to three trials and 12 times in each play
the subject is asked the question
``How happy are you at this moment?''

We model this experiment as a reinforcement learning problem with
$150$ different states representing
the different trials that the subject can see
(the possible combinations of certain rewards and gamble outcomes),
and two actions \texttt{certain} and \texttt{gamble}.
For example,
in a trial the subject could be presented with the choice between
a certain reward of 30 and a 50--50 gamble between 0 and 72.

The expected reward after each state is uniquely determined by
the state's description,
but the agent has subjective uncertainty about the
value of the state, since it does not know
which states (types of trials) will follow the current one
or how these states are distributed.
(In the experiment,
the a priori expected outcome of a trial (the average value of the states) was
$5.5$ points,
and the maximum gain or loss in a single trial is $220$ points.)
Furthermore, the subject might incorrectly estimate the value of the gamble:
although the visualisation correctly suggests that
each of the two outcomes is equally likely,
the subject might be uncertain about this,
or simply compute the average incorrectly.

Rutledge et al.\ model happiness as an affine-linear combination of
the certain reward (CR),
the expected value of the gamble (EV)
and the reward prediction error at the outcome of the gamble (RPE).
The weights for this affine-linear combination
were learned through linear regression
on fMRI data from another similar experiment on humans
($w_{\text{CR}} = 0.52$, $w_{\text{EV}} = 0.35$, and $w_{\text{RPE}} = 0.8$
for $z$-scored happiness ratings).

The data was kindly provided to us by Rutledge.
We disregard the first happiness rating that occurs before the first trial.
Moreover, we removed all $762$ subjects
whose happiness ratings (other than the first) had a standard deviation of $0$.

To test our happiness definition on the data,
we have to model how humans estimate the value of a history.
We chose a very simple model where a subject's expectation
before each trial is the average of the previous outcomes.
We use the empirical distribution function,
that estimates all future rewards as the average outcome so far:
\[
\rho(r_n \mid h) := \frac{\#\{ k \leq t \mid r_k = r_n \}}{t}
\text{ for $n > t$}.
\]
With \autoref{eq:subjective-expectation},
this gives a value function estimate of
$\hat{V}(a_1 o_1 r_1 \ldots a_t o_t r_t)
= \tfrac{1}{t (1 - \gamma)} \sum_{i=1}^t r_i$.

We assume that the subject typically
computes the expected value of the trial correctly, i.e.\
$E[r_t \mid h] := \max \{ \text{CR}, \text{EV} \}$.
We calculate happiness at time $t$ with the formula from
\autoref{prop:happiness-as-subjective-expectation}.

\begin{figure}[t]
\begin{center}
\includegraphics[width=0.8\textwidth]{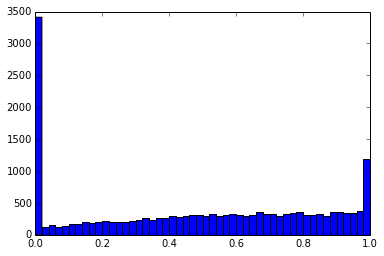}
\end{center}
\caption{
Distribution of the discount factor $\gamma$
(for happiness aggregation and rewards)
fit per subject ($N = 17658$).
The plot for Rutledge et al.'s model looks similar.
}
\label{fig:gamma-distribution}
\end{figure}

Since subjects' happiness ratings are not given on every trial,
we use geometric discounting with discount factor $\gamma$
to aggregate happiness from previous trials:
\[
\text{predicted happiness}
:= \sum_{k=1}^t \gamma^{t-k} \happy(h_k, \hat{V}_k),
\]
where $h_k$ is the history at time step $k$ and
$\hat{V}_k$ is the value function estimate at time step $k$.
For each subject,
we optimise $\gamma$ such that the Pearson correlation $r$ is maximised.
The result is that the majority of subjects
use a $\gamma$ of either very close to $0$ or very close to $1$
(see \autoref{fig:gamma-distribution}).
A possible explanation of this phenomenon could be that
it is unclear to the human subjects whether
they are to indicate their instantaneous or their cumulative happiness.

We use the \emph{sample Pearson correlation $r$}, its square $r^2$ and the
coefficient of determination $R^2$ to evaluate each model.
For $n$ data points
(in our case $n = 11$, the number of happiness ratings per subject),
the sample Pearson correlation $r$ is defined as
\[
r = r_{xy} = \frac{ \sum_{i=1}^n  (x_i - \bar{x}) (y_i - \bar{y}) }
 {\sqrt{\sum_{i=1}^n (x_i - \bar{x})^2 \sum_{i=1}^n (y_i - \bar{y})^2}}
\]
where $\bar{x}=\frac{1}{n}\sum_{i=1}^n x_i$ is the sample mean.
The coefficient of determination $R^2$ on $n$ data points
with (training) data $x \in X$ and predictions $y\in Y$ is
\[
R^2 = 1 - \frac{\sum_{i=1}^n (y_i - x_i)^2} {\sum_{i=1}^n (x_i - \bar{x})^2}
\]
We calculate $R^2$ only on z-scored terms;
z-scoring does not affect $r$.

Our model's predicted happiness
correlates fairly well with reported happiness
(mean $r = 0.56$, median $r^2 = 0.41$, median $R^2 = 0.27$)
while fitting individual discount factors,
comparative to Rutledge et al.'s model
(mean $r = 0.60$, median $r^2 = 0.47$, median $R^2 = 0.36$) and
a happiness=cumulative reward model
(mean $r = 0.59$, median $r^2 = 0.46$, median $R^2 = 0.35$).
So this analysis is inconclusive.
\autoref{fig:correlation-our-model} and \autoref{fig:correlation-Rutledge}
show the distribution of the correlation coefficients.
We emphasise that our model was derived a priori and then tested on the data,
while their model has three parameters (weights for CR, EV, and RPE)
that were fitted on data from humans using linear regression.

\begin{figure}[t]
\centering
\begin{minipage}{.47\textwidth}
\begin{center}
\includegraphics[width=\textwidth]{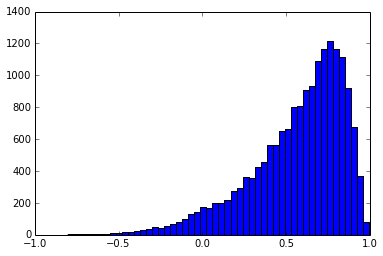}
\end{center}
\captionof{figure}{
	The distribution of the correlation coefficients for our happiness model
	($N = 17658$).
}
\label{fig:correlation-our-model}
\end{minipage}~~~~
\begin{minipage}{.47\textwidth}
\begin{center}
\includegraphics[width=\textwidth]{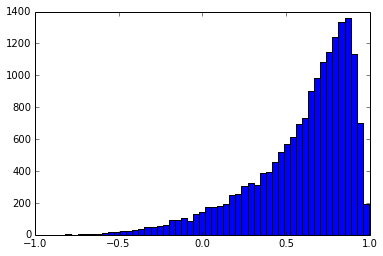}
\end{center}
\captionof{figure}{
	The distribution of the correlation coefficients for Rutledge et al.'s model
	($N = 17658$).
}
\label{fig:correlation-Rutledge}
\end{minipage}
\end{figure}

It is well known that humans do not value rewards linearly.
In particular, people appear to weigh losses more
heavily than gains, an effect known as \emph{loss aversion}.
To model this, we set
\[
\text{reward} :=
\begin{cases}
\text{outcome}^\alpha &\text{if outcome} > 0 \\
-\lambda \cdot (-\text{outcome})^\beta &\text{otherwise}.
\end{cases}
\]
Using parameters found by Rutledge et al.\
(mean $\lambda = 1.7$, mean $\alpha = 1.05$, mean $\beta = 1.01$)
we also get a slightly improved agreement
(mean $r = 0.60$, median $r^2 = 0.48$, median $R^2 = 0.39$).
\autoref{tab:results} lists these results.

\begin{table}
\begin{center}
\begin{filecontents*}{results.dat}
Model, gamma, r, Medianr2, R2, MedianR2
normal, $0.52 \pm 0.34$ , $0.56 \pm 0.30$ , 0.41, $0.12 \pm 0.61$ , 0.27
loss aversion, $0.53 \pm 0.34$ , $0.60 \pm 0.31$ , 0.48, $0.20 \pm 0.61$ , 0.39
Rutledge et al., $0.55 \pm 0.35$ , $0.60 \pm 0.30$ , 0.47, $0.19 \pm 0.60$ , 0.36
pleasure, $0.55 \pm 0.35$, $0.59 \pm 0.30$, 0.46, $0.18 \pm 0.60$, 0.35
\end{filecontents*}
\pgfplotstabletypeset[
every head row/.style={before row=\hline, after row=\hline\hline},
every last row/.style={after row=\hline},
every row no 1/.style={after row=\hline\hline},
every row no 2/.style={after row=\hline\hline},
every first column/.style={
column type/.add={|}{}},
every last column/.style={
column type/.add={}{|}}, col sep=comma,
columns/Model/.style={string type},
columns/gamma/.style={string type, column name=$\gamma$},
columns/r/.style={string type, column name=$r$},
columns/Medianr2/.style={string type, column name=med $r^2$},
columns/R2/.style={string type, column name=$R^2$},
columns/MedianR2/.style={string type, column name=med $R^2$},
]{results.dat}
\end{center}
\caption{Discount factor $\gamma$,
correlation coefficient $r$, and goodness-of-fit measure $R^2$
for various versions of our model, compared to Rutledge et al.\
and a happiness=pleasure model
(N = 17658).}
\label{tab:results}
\end{table}

\fi 

\end{document}